\documentclass{ieeeaccess}
\usepackage{cite}
\usepackage{amsmath,amssymb,amsfonts}
\usepackage{algorithmic}
\usepackage{graphicx}
\usepackage{textcomp}
\usepackage{multirow}
\usepackage{float}
\usepackage{placeins}

\def\BibTeX{{\rm B\kern-.05em{\sc i\kern-.025em b}\kern-.08em
    T\kern-.1667em\lower.7ex\hbox{E}\kern-.125emX}}
\begin{document}
\history{Date of publication January 25, 2022.}
\doi{10.1109/ACCESS.2022.3146059}

\title{Prototype Memory for Large-scale\\ Face Representation Learning}

\author{\uppercase{Evgeny Smirnov}\authorrefmark{1}, 
\uppercase{Nikita Garaev}\authorrefmark{1}, \uppercase{Vasiliy Galyuk}\authorrefmark{1}, and \uppercase{Evgeny Lukyanets}\authorrefmark{2},}
\address[1]{Speech Technology Center, 194044 Saint Petersburg, Russia}
\address[2]{Department of Information Technologies and Programming, ITMO University, 197101 Saint Petersburg, Russia}

\tfootnote{This work was supported by the Analytical Center for the Government of the Russian Federation (IGK 000000D730321P5Q0002), agreement No. 70-2021-00141.}

\corresp{Corresponding author: Evgeny Smirnov (e-mail: smirnov-e@speechpro.com).}

\begin{abstract}
Face representation learning using datasets with a massive number of identities requires appropriate training methods. Softmax-based approach, currently the state-of-the-art in face recognition, in its usual "full softmax" form is not suitable for datasets with millions of persons. Several methods, based on the "sampled softmax" approach, were proposed to remove this limitation. These methods, however, have a set of disadvantages. One of them is a problem of "prototype obsolescence": classifier weights (prototypes) of the rarely sampled classes receive too scarce gradients and become outdated and detached from the current encoder state, resulting in incorrect training signals. This problem is especially serious in ultra-large-scale datasets. In this paper, we propose a novel face representation learning model called Prototype Memory, which alleviates this problem and allows training on a dataset of any size. Prototype Memory consists of the limited-size memory module for storing recent class prototypes and employs a set of algorithms to update it in appropriate way. New class prototypes are generated on the fly using exemplar embeddings in the current mini-batch. These prototypes are enqueued to the memory and used in a role of classifier weights for softmax classification-based training. To prevent obsolescence and keep the memory in close connection with the encoder, prototypes are regularly refreshed, and oldest ones are dequeued and disposed of. Prototype Memory is computationally efficient and independent of dataset size. It can be used with various loss functions, hard example mining algorithms and encoder architectures. We prove the effectiveness of the proposed model by extensive experiments on popular face recognition benchmarks.
\end{abstract}

\begin{keywords}
Deep neural networks, face recognition, representation learning, softmax acceleration
\end{keywords}

\titlepgskip=-15pt

\maketitle

\section{Introduction}
\label{sec:introduction}
\PARstart{F}{ace} recognition is one of the most established technologies \cite{dfrsurvey, du2020elements} of computer vision. With the help of modern deep neural architectures \cite{hinton2015deep} and large-scale training datasets \cite{guo2016ms, an2020partial}, current face recognition models \cite{wang2018cosface, deng2019arcface, deng2020sub, huang2020information} perform at superhuman levels \cite{taigman2014deepface} on the million-scale face identification and verification tasks \cite{kemelmacher2016megaface}. However, there are still a number of open problems in face recognition, such as large pose and age variations \cite{cfpfp, cplfw, agedb, calfw}, racial bias \cite{RFW}, unconstrained face image conditions \cite{ijbc}, usage of heavy makeup, disguise \cite{singh2019recognizing}, and face masks \cite{damer2021masked}.

The most straightforward way to combat these problems is a construction of large and (preferably) balanced training datasets with sufficient diversity of face image variations. Indeed, the best performing models in popular face recognition benchmarks are trained on a datasets, containing hundreds of thousands \cite{an2020partial, deng2020sub, Zhang_2020_CVPR} or even millions \cite{schroff2015facenet, zhu2021webface260m} of persons.

Training on the datasets of a million-person scale provides a set of challenges. One of them is the choice of training method. First possible way to train a model on a dataset with a large number of persons is to rely on a triplet-based training scheme \cite{schroff2015facenet}. It is scalable, but usually slow and less accurate than alternatives like softmax-based methods \cite{liu2017sphereface, deng2019arcface, wang2018cosface}.

Softmax-based methods consider face recognition as a classification task, where persons are classes, and the model is trained to classify an image as belonging to one of the classes in the training set. These methods are proven to perform well, but they have a problem with scaling. Computation and GPU memory requirements of these methods are dependent on a number of classes in the dataset. Small and medium-sized \cite{cao2018vggface2, bansal2017umdfaces, wang2018devil} datasets with less than $100,000$ classes could be used for softmax-based methods efficiently, but training on the datasets with millions of classes becomes too computationally demanding or even infeasible.

\begin{figure*} [h]
	\begin{center}
		\includegraphics[width=1.0\linewidth]{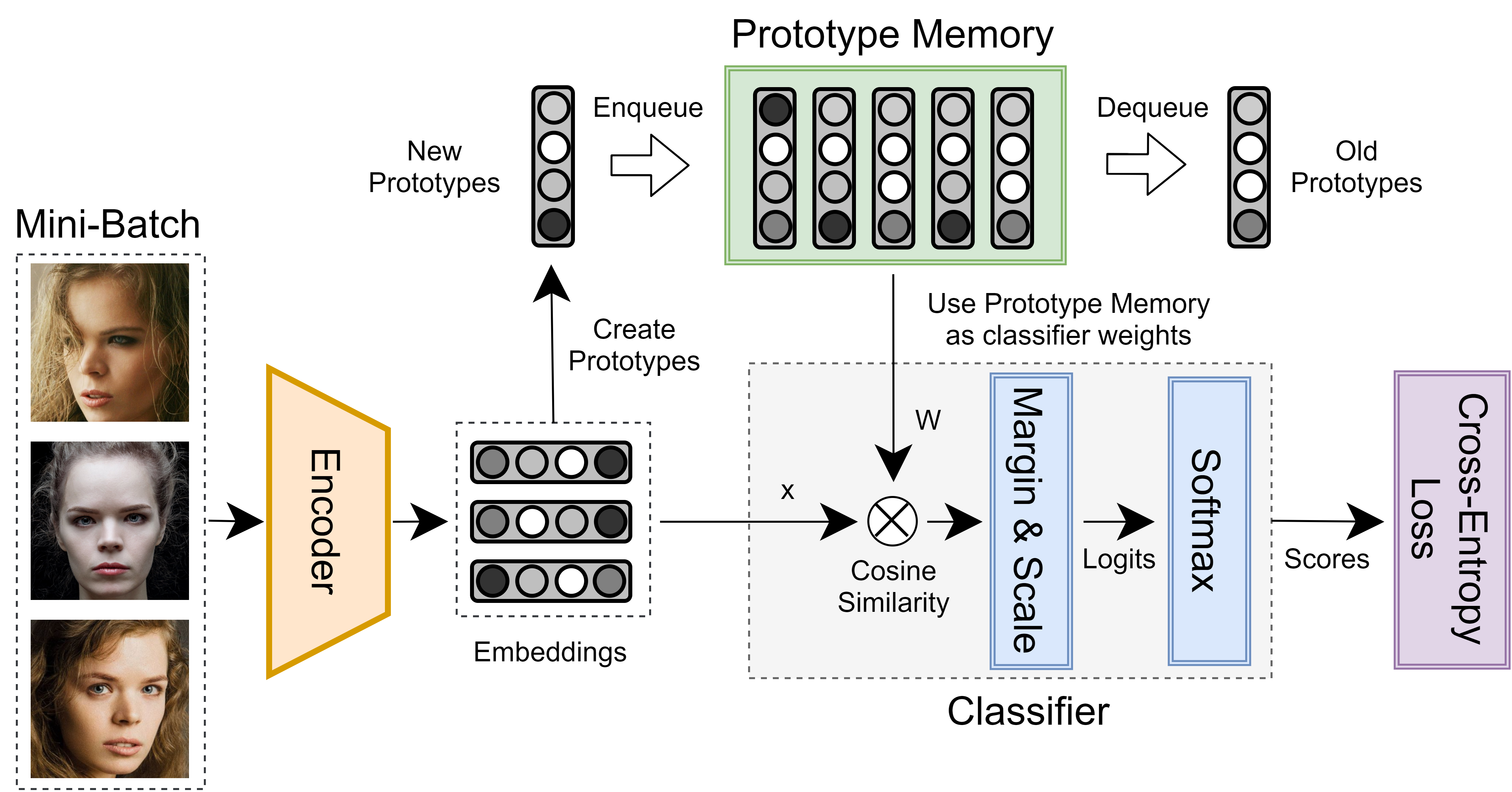}
	\end{center}
	\caption{Scheme of the face representation learning with Prototype Memory. Mini-batch with several images per class is passed through the encoder, producing exemplar embeddings. Same-class embeddings are used to generate new class prototypes. These prototypes are enqueued to the memory module and used as the classifier weights in softmax classifier-based training. When the memory is full, the oldest prototypes are dequeued and disposed of.}
	\label{fig:Smirn1}
\end{figure*}

To overcome the difficulties of large dataset softmax-based training, several methods were proposed \cite{zhang2018accelerated, he2020softmax, an2020partial, liu2019adaptiveface}. These methods are based on the "sampled softmax" approach, where only a small subset of the total set of classes is used in the current training iteration. The advantage of this approach is that there is much less computation and GPU memory involved compared to the "full softmax" training. Only selected class prototypes (classifier weights) are moved to the GPU memory to perform training iteration. All other prototypes stay in RAM or in other large-capacity memory. Whereas the "full softmax" approach has the advantage of more complete information available at the loss computation stage, comparable results could be achieved with "sampled softmax" using smart selection of sampled classes \cite{zhang2018accelerated, dm, liu2019adaptiveface} and large enough training datasets \cite{an2020partial}.

"Sampled softmax"-based methods are good solutions for some problems in face recognition model training on large-scale datasets, but they also have a set of disadvantages. First, they still need to store prototypes of all classes in memory. For multi-million-scale datasets and large prototype size it could become an unnecessary bottleneck. Also it means that the dataset must be fixed during the training process, and no new classes could be added. Second, operations of sampled prototype transfer from RAM to GPU memory and back add extra computational overhead. Third, class prototypes, which are not sampled in the current iteration, do not get gradients, and over time become detached from the encoder and begin to represent their class incorrectly in the embedding space. It leads to inaccurate training signals and subsequent performance degradation. We refer to this situation as a problem of "prototype obsolescence". It is even more dangerous in large-scale datasets, where the frequency of sampling each individual class is very low.

In this paper, we propose Prototype Memory - a novel face representation learning model, suitable for training on a dataset, regardless of the number of identities in it. Prototype Memory is based on the idea of "online" class prototype generation. New class prototypes are approximated using groups of same-class exemplar embeddings in the current mini-batch, and stored in a memory of limited size. Prototypes in memory are used in a role of classifier weights to perform softmax-based training. Prototypes are updated using gradients and also refreshed with recent exemplar embeddings, when they are available. The memory is continuously filled with prototypes until it reaches its full capacity. In this case, to free the memory for new prototypes, the oldest ones are removed from the memory and disposed of.

Prototype Memory shares the advantages of other "sampled softmax"-based methods, but is free from many of their disadvantages. First, there is no need to store all class prototypes, only those in current Prototype Memory. This behavior allows us to use datasets with an unlimited number of persons and train in class-incremental style. Second, the proposed approach has no overhead on RAM to GPU memory transfer operations: all Prototype Memory operations can be performed entirely on GPU. Third, class prototypes, used for training, are continuously updated and kept up-to-date by means of gradients and refreshing algorithm. Outdated prototypes are disposed of. This way we ensure that class prototypes are closely connected to the current encoder state and correctly represent their corresponding classes in the embedding space. It helps to solve the problem of "prototype obsolescence".

Prototype Memory is orthogonal to loss functions, neural network architectures, hard example mining methods, etc. It can be combined with many different methods, including smart class \cite{dm, zhang2018accelerated, liu2019adaptiveface} and example \cite{ae, liu2019adaptiveface} sampling algorithms, margin-enhanced softmax variants \cite{curricularface, deng2019arcface, zeng2020npcface} and other state-of-the-art face recognition methods \cite{kim2020broadface, groupface, meng2021magface, liu2021interclass, xu2021searching}.

We summarize the contributions as follows:

\begin{itemize}
	\item We propose Prototype Memory model for face representation learning, which can be used to efficiently train face recognition models on very large datasets.
	\item We prove the effectiveness of the proposed model with extensive experiments on different face recognition benchmarks.
	\item We present algorithms of hard example mining and knowledge distillation, suitable for Prototype Memory.
\end{itemize}

\section{Related Work}

\subsection{Face Recognition}
Current state-of-the-art methods in the area of face recognition are based on the training of deep neural networks on large face image datasets \cite{parkhi2015deep, schroff2015facenet}. The usual pipeline of face recognition is to detect faces in image with a face detector \cite{7553523, deng2020retinaface}, align and crop them, and then pass through the encoder (deep neural network) to get face representations in a form of $L_2$-normalized vectors (embeddings). Encoder is trained to represent different face images of the same person close to each other in the embedding space (according to the cosine similarity measure), and face images of two different persons - far from each other. There are two main ways to train an encoder to perform this task. First one is to group images in pairs \cite{chopra2005learning} or triplets \cite{schroff2015facenet} and use their embeddings to compute gradients, pushing them to each other - for the same-class pairs, and apart from each other - for the pairs of different classes. 

Second way to train proper face encoders is to present a task of face recognition in the form of softmax-based image classification \cite{taigman2014deepface} and apply additional constraints to the resulting image representations \cite{wen2016discriminative, ranjan2017l2} to ensure they possess necessary properties. Classifier weights in this approach could be viewed as class representatives (prototypes) in the embedding space, and exemplar embeddings are trained to move closer to their corresponding prototypes and away from the prototypes of other classes. With proper normalization, prototypes bear conceptual similarity to the class centers \cite{wang2017normface} in the embedding space. This approach is currently considered state-of-the-art and represented with such methods as SphereFace \cite{liu2017sphereface}, ArcFace \cite{deng2019arcface}, CosFace \cite{wang2018cosface}, CurricularFace \cite{curricularface} and many others \cite{wang2018additive, wang2020mis, zeng2020npcface, duan2019uniformface, zhao2019regularface, liu2019adaptiveface}.

There are also many complementary methods proposed to build better face recognition models by promoting desired properties of the produced face representations, such as robustness to noisy labels \cite{deng2020sub} and low image resolution \cite{khalid2020resolution}, invariance to age \cite{zhao2020towards} and pose \cite{he2020deformable}, ability to mitigate racial bias \cite{wang2020mitigating} and domain imbalance \cite{cao2020domain, groupface}, to improve the fairness of representations \cite{Liu_2019_ICCV}. There are also methods, proposed to overcome the problems with the situations of difficult face appearance variations, like deliberately disguised faces \cite{cmb, ae} or faces in medical masks \cite{boutros2021unmasking}. 

All these methods are used to improve face recognition models, but the main key of success is the usage of large training datasets. For example, datasets, used to train state-of-the-art models in MegaFace challenge \cite{kemelmacher2016megaface}, are composed of hundreds of thousands \cite{Zhang_2020_CVPR} and even millions \cite{zhu2021webface260m} of persons. To use this kind of datasets with the softmax-based approach there is a need of some kind of softmax acceleration methods.

\subsection{Softmax Acceleration}
Utilization of softmax function is currently the main approach to perform classification tasks in the field of machine learning. In case of small number of classes, softmax is very fast, but when the number of classes reaches millions, softmax computation becomes a bottleneck. Recent works propose a number of ways to accelerate the computation of softmax-based classification. One line of research is looking for the opportunities to accelerate softmax calculation with more efficient algorithms \cite{milakov2018online, dukhan2020two, bamler2020extreme} and parallelization \cite{deng2019arcface, an2020partial}, the other one is trying to approximate softmax \cite{shim2017svd, joulin2017efficient}, split it into several parts \cite{wu2016deep}, use softmax hierarchically \cite{wydmuch2018no, shen2017self} or sample classes stochastically \cite{reddi2019stochastic, NEURIPS2019_e43739bb, blanc2018adaptive, yu2017tapas}.

In the area of face recognition, the latter approach is the most promising. For example, a variant of "sampled softmax" approach, Selective Softmax \cite{zhang2018accelerated} reduces the computation cost and memory demand of softmax computation by selecting only a subset of "active classes" for each mini-batch based on a set of dynamic class hierarchies, constructed on-the-fly. D-Softmax-K \cite{he2020softmax} randomly samples a subset of negative classes and uses two different loss functions for intra-class and inter-class training. Random Prototype Softmax and Dominant Prototype Softmax, proposed in \cite{zhu2018large}, store a matrix of prototypes in the non-GPU memory and use different ways to select and move prototypes to the GPU memory to perform training in the bisample face recognition scenario. Hard Prototype Mining \cite{liu2019adaptiveface} is used to adaptively select a small number of hard prototypes, most similar to the samples in the mini-batch in each iteration. Only these prototypes are used in softmax calculation, thus accelerating the training.

Positive Plus Randomly Negative (PPRN) \cite{an2020partial} is another softmax acceleration method. Its core idea is to keep a full classifier weight matrix in RAM and sample positive classes and a subset of random negative classes at each iteration to perform training. Provided with a large-scale training dataset, this method of class sampling achieves the results, comparable to the "full softmax"-based training methods.

\subsection{Weight Imprinting}
In the classic softmax-based training, classifier weights for all training classes are initialized with random values in the beginning of the learning process. However, with this kind of initialization all classes are need to be determined before the training, and their set cannot be altered after that. To achieve the possibility of inserting new classes to the trained network, \cite{qi2018low} proposed a method of weight imprinting. With this method, embedding vectors of the examples of the new class can be used to initialize weights for this class, thereby extending the classifier. To generate a weight for a new class, embeddings of its examples are averaged, and the resulting vector is $L_2$-normalized and inserted into the classifier. Another closely related model is Prototypical Networks \cite{snell2017prototypical}. They were proposed for the task of few-shot learning and use a support set of several examples per class to approximate the prototype of this class. Prototypes are calculated in the embedding space using the mean embedding of the support set. Classification is then performed by finding the nearest class prototype. There are a number of ways to use and adapt prototype-based learning in different areas \cite{zhangrbf, li2021prototypical, de2021continual}. In case of face recognition, class prototypes could be calculated using averaged and $L_2$-normalized face image embeddings, belonging to the same class.

\subsection{Exemplar Memory}
Exemplar memory is a way to improve the neural network training by keeping examples or their embeddings in memory between iterations and use them to get better training information. Exemplar memory was used in \cite{zhong2019invariance} to store the features of target domain and enforce invariance constraints for the task of person re-identification. Another proposed variant of exemplar memory is called Cross-Batch Memory (XBM) \cite{wang2020cross}. It was used to improve embedding learning by keeping embeddings from past iterations and using them to find better hard examples and learn better embeddings. Memory-based Jitter \cite{liu2020memory} enhances intra-class diversity for the tail classes in the long-tail learning problem by keeping in a memory bank different versions of exemplar features from previous iterations. BroadFace \cite{kim2020broadface} is another variant of exemplar memory, proposed for face recognition. BroadFace stores a large number of exemplar embeddings from previous iterations and uses them to compute better training signals for the classifier weight matrix updates. Since the exemplar embeddings are computed with different versions of encoder from different training steps, they are suffering from the problem of obsolescence. To prevent it, authors proposed compensation function: using the difference between prototypes at current iteration and at the iteration, when the embedding was enqueued to the memory, calculate the compensation value and apply it to the embedding. Compensation function is helpful for embedding obsolescence, but it cannot be applied to prevent prototype obsolescence, because in contrast with the BroadFace setting, in the "sampled softmax" situation there is no "actual" version of each class prototype at each iteration to get the compensation value from.

\section{Proposed Method}

In this section we propose Prototype Memory model for face representation learning. It combines the advantages of weight imprinting (online generation of class prototypes using groups of class exemplars), exemplar memory (keeping useful information between training iterations) and accelerated softmax methods (faster and more memory-efficient than "full softmax") and could be used to train face recognition architectures on datasets of any size without the problem of prototype obsolescence.

\subsection{Motivation} 

Current state-of-the-art face recognition models \cite{deng2019arcface} use softmax-based training strategy. When the dataset is very large (millions of persons), this type of strategy becomes too memory and computationally expensive: for the training with "full softmax", all classifier weights should be placed in the GPU memory, and for each training example all classes in the classifier are needed to compute the loss function and gradients.

Current "sampled softmax"-based methods \cite{zhang2018accelerated, he2020softmax, an2020partial, liu2019adaptiveface}, while solving some of the problems of "full softmax", still have the need to keep all classifier weights somewhere, and suffer from the problem of "prototype obsolescence" - situations when the class prototype has not received gradients for so long time, that it does not represent its class correctly anymore. This problem is unavoidable in current methods if just a subset of classifier weights is updated at each iteration.

One more problem we want to solve is the inability of the current face recognition methods to use training datasets in class-incremental fashion, when the dataset could be updated with new classes at any time, without training problems.

\subsection{Prototype Memory}

The scheme of face representation learning with the proposed Prototype Memory model is presented in Fig. 1. Prototype Memory includes a memory module and algorithms for prototype generation, prototype refreshing and prototype disposal.

At each training iteration, several images of some class are passed through the encoder network, producing $L_2$-normalized embeddings. These embeddings are used to generate a new prototype for the corresponding class. New class prototypes are enqueued to the Prototype Memory module. Prototypes in this module are later used as the classifier weights to perform usual softmax-based face recognition model training with some appropriate loss function.

Prototypes in the Prototype Memory are updated with the gradients. If the class, currently already placed in the Prototype Memory, gets new examples in the current mini-batch, it also uses their embeddings to "refresh" its prototype and move it to the start of the queue. When the Prototype Memory is filled to its capacity, oldest prototypes, which haven't been refreshed with exemplar embeddings for many iterations and became outdated, are dequeued and disposed of to free the memory slots for the new ones.

Prototype Memory model has a set of hyperparameters:

\begin{itemize}
	\item \textbf{Prototype Size (D)} - it is the size of the prototype vectors. This size must be the same as the size of the embedding vectors, produced by the encoder.
	\item \textbf{Memory Size (M)} - is the number of prototypes, which could be placed in the memory module.
	\item \textbf{Number of images per class (k)}, used for the prototype generation.
	\item \textbf{Refresh Ratio (r)} - this hyperparameter is responsible for the strength of the prototype refreshing.
\end{itemize}

More detailed description of the Prototype Memory algorithms is presented below and on Fig. 3.

\subsubsection{Prototype Generation}
Prototype generation scheme is presented at Fig. 3a. Mini-batch, containing the images of several classes, with the fixed number of images per class (set by a hyperparameter $k$), is passed through the encoder network. Resulting embeddings for each person (identity) are averaged, and the average embedding is $L_2$-normalized:
\begin{equation}
\mathbf{P}_{new} = F_{norm}(\frac{\sum_{j=1}^{k} \mathbf{x}_{j}}{k}),
\end{equation}
$\mathbf{x}_{j}$ is the $j$-th exemplar embedding, produced by encoder, $k$ is the number of images, used for prototype generation, and $F_{norm}$ is $L_2$-normalization function. The resulting vector $\mathbf{P}_{new}$ is the new class prototype. New prototypes are calculated for each class in the mini-batch, and then enqueued to the Prototype Memory (Fig. 3b). 

\subsubsection{Prototype Update and Refreshing}
Prototype Memory module has fixed size, set by a hyperparameter $M$. New prototypes are enqueued to the Prototype Memory (and later dequeued from it) in order of appearance in training. When the prototypes are in Prototype Memory, they are used in a role of the classifier weights for usual softmax classifier-based training with a margin-based face-recognition-specific loss function like ArcFace \cite{deng2019arcface}. These prototypes receive gradients and are updated according to them.

When the class has new examples in the current mini-batch, and already has a prototype in the Prototype Memory, additionally to the usual gradient-based update, it is also "refreshed" using current mini-batch embeddings with:
\begin{equation}
\mathbf{P}_{upd} = F_{norm}(r * \mathbf{P}_{new} + (1 - r) * \mathbf{P}_{mem}),
\end{equation}
where $\mathbf{P}_{upd}$ is updated prototype, $\mathbf{P}_{mem}$ is a prototype from memory, $\mathbf{P}_{new}$ is a prototype, calculated with the embeddings in the current mini-batch, $F_{norm}$ means $L_2$-normalization, and $r$ is a hyperparameter (refresh ratio). Prototype refreshing is performed to force the prototypes in Prototype Memory to be up-to-date to the current encoder state. When the prototype is refreshed, it is moved to the start of the queue as if it is a new prototype.

\subsubsection{Prototype Disposal}
Prototype disposal scheme is presented at Fig. 3b. When the Prototype Memory is full and has no free slots for new prototypes, there is a need to dispose of the oldest prototypes. These are the prototypes, which resided in the memory for the longest time, received no refreshing from their examples for a while and became outdated. Most likely, the "real" prototypes of these classes, if computed using the current encoder state and the images of this class, would be far from these prototypes in memory. To free the memory space for the new prototypes, these prototypes are dequeued and disposed of. Freed memory slots are then used for new prototypes.

\subsection{Solution to prototype obsolescence}

Prototype obsolescence problem of "sampled softmax"-based training methods is illustrated in Fig. 2a. Left part of the figure demonstrates a situation, when the class prototype is up-to-date, and the distance between prototype and class center (mean embedding for all class examples) is small. Right part of the figure shows a situation after a number of iterations, during which this class has not been sampled, and the prototype was not updated. Encoder has evolved, and now it produces exemplar embeddings in different points of the hypersphere. This resulted in a new position of the class center. Class prototype, however, stayed in the same place, and now the distance between prototype and class center became large. This prototype became obsolete, it is not representing the class correctly in the embedding space anymore.

Fig. 2b illustrates a similar situation, but for Prototype Memory. Instead of using the same old prototype from previous iterations, a new prototype is generated using exemplar embeddings in the current mini-batch. In this way it is kept up-to-date, even after a large number of iterations between the mini-batches, when the same class is sampled. Since new class prototypes are generated using up-to-date encoder, the distance between class centers and prototypes stays small, so the problem of prototype obsolescence is solved.

\begin{figure}[h]
	\begin{center}
		\includegraphics[width=1.0\linewidth]{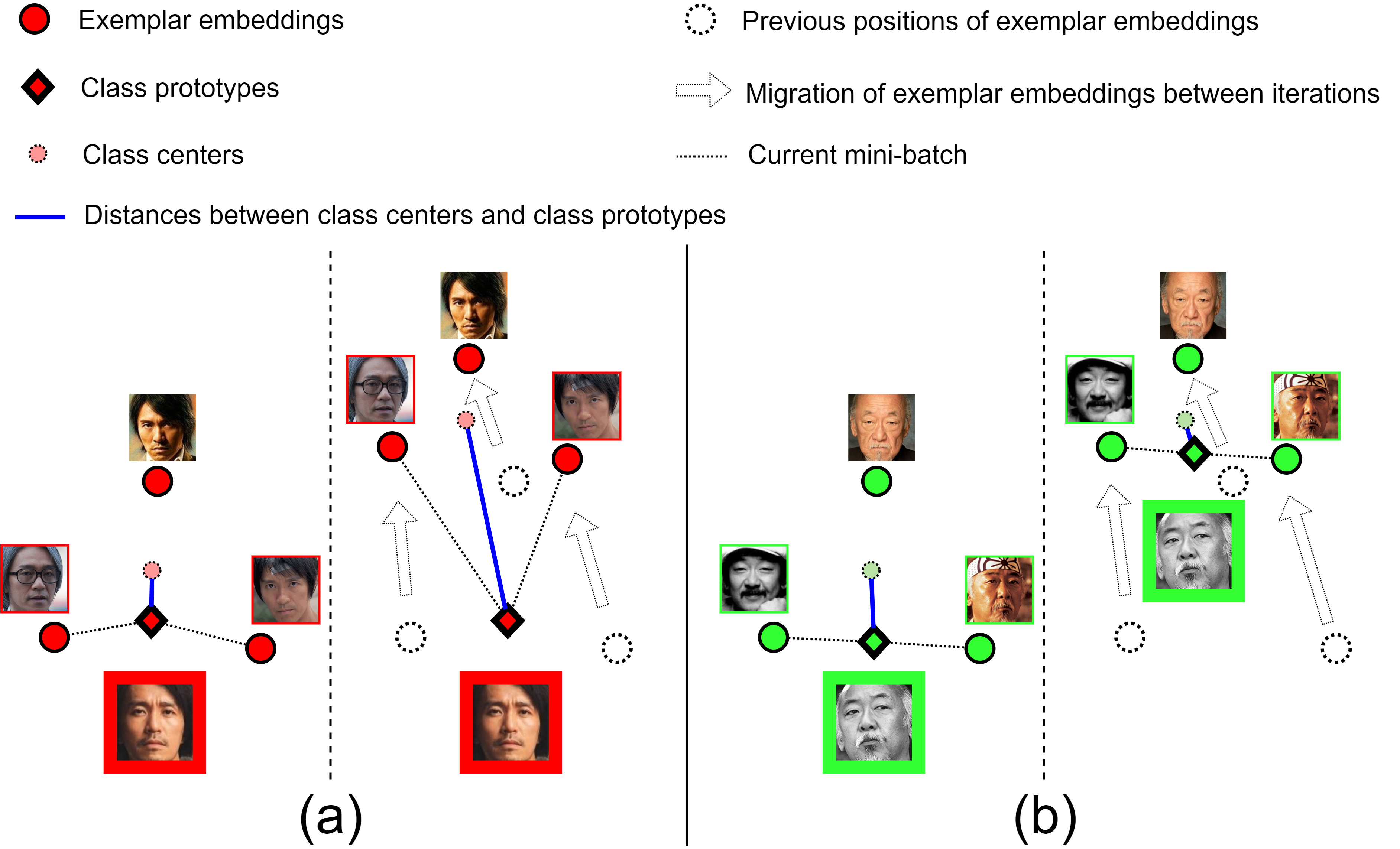}
	\end{center}
	\caption{(a) Illustration of the prototype obsolescence problem in the "sampled softmax"-based training methods, (b) Illustration of the absence of prototype obsolescence problem in Prototype Memory}
	\label{fig:Smirn2}
\end{figure}

\begin{figure*}[h]
	\begin{center}
		%		\fbox{\rule{0pt}{2in} \rule{.9\linewidth}{0pt}}
		\includegraphics[width=1.0\linewidth]{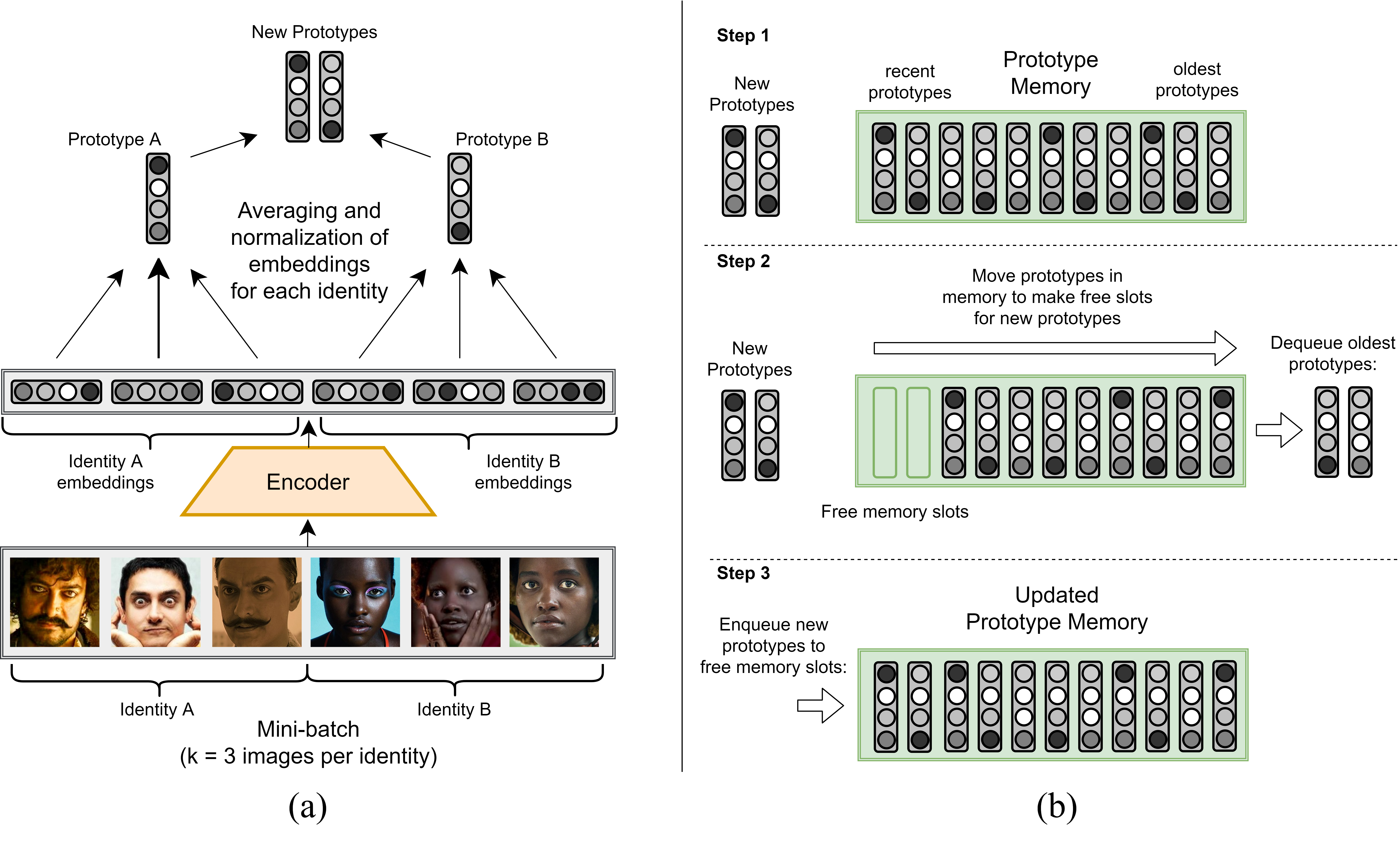}
	\end{center}
	\caption{a) Scheme of prototype generation, b) Scheme of Prototype Memory update with new prototypes and the disposal of old prototypes}
	\label{fig:Smirn3}
\end{figure*} 

\subsection{Training Specifics}

\subsubsection{Group-based Mini-batch Sampling}
For the prototype generation stage, to create prototypes closer to the corresponding class centers, we need to ensure that each class in the mini-batch has some minimum number of images $k$, sampled together. So, we need to sample images into the mini-batches accordingly. In this paper we utilize two different types of methods, used to sample mini-batches with groups of at least $k$ images for each class:

\begin{itemize}
	\item \textbf{Group-based iterate-and-shuffle}: First, images in the training dataset are randomly combined into the single-class groups of the same size $k$. Then these groups are shuffled and iteratively sampled to the mini-batch. When all groups are used, the process is repeated. With this kind of mini-batch sampling, we ensure that all images in the dataset will be sampled approximately equal number of times.
	\item \textbf{Group-based classes-then-images}: First, a number of classes are sampled from the training dataset. Then for each class a group of $k$ random images sampled to the current mini-batch. With this kind of mini-batch sampling, we ensure that all classes in the dataset will be sampled approximately equal number of times. 
\end{itemize}

These methods could be combined with algorithms of hard class mining \cite{dm} and hard example mining \cite{ae}, or used together as parts of a composite mini-batch \cite{cmb}.

\subsubsection{Loss Functions}
Prototype Memory is a model, independent of the choice of the loss function \cite{wang2020comprehensive}. Most loss functions, suitable for usual softmax-based face representation learning, could be also used to perform training with Prototype Memory. In our experiments, we use Large Margin Cosine Loss \cite{wang2018cosface}:

\begin{equation}
L = -\frac{1}{N}\sum_{i=1}^{N}\log \frac{e^{s(\cos\theta_{y_{i}} - m)}}{e^{s(\cos\theta_{y_{i}} - m)} + \sum_{j=1,j \neq y_{i}}^{M} e^{s\cos\theta_{j}}},
\end{equation}

where $N$ is a size of the mini-batch, $M$ is the size of the memory module, $y_{i}$ is the ground-truth class, $\theta_{j}$ is an angle between $i$-th exemplar embedding and $j$-th class prototype, $s$ is a scale parameter, and $m$ is margin.

We also performed experiments with D-Softmax \cite{he2020softmax} loss:

\begin{equation}
	L_D = \frac{1}{N}\sum_{i=1}^{N}\log(1+\frac{\epsilon}{e^{s\cos\theta_{y_{i}}}})+\log(1+\sum_{j \neq y_{i}}^{M} e^{s\cos\theta_{j}}),
\end{equation}

where $\epsilon = e^{ds}$, $d$ is a hyperparameter (optimization termination point), $s$ is a scale parameter, $N$ is a size of the mini-batch, $M$ is the size of the memory module, $y_{i}$ is the ground-truth class, and $\theta_{j}$ is an angle between $i$-th exemplar embedding and $j$-th class prototype. Other similar loss functions \cite{deng2019arcface, wang2020mis, curricularface} are applicable too.

\subsection{Hard Example Mining for Prototype Memory}
Prototype Memory could be used together with hard example mining \cite{ae} and hard class mining methods like Doppelganger Mining \cite{dm}, DP-Softmax \cite{zhu2018large} or HPM \cite{liu2019adaptiveface}. Careful selection of hard classes and examples is used to decrease the number of required training iterations and to achieve better performance in difficult face recognition scenarios. Most methods could be used for Prototype Memory without modifications, but some of them need to be adapted to the "sampled softmax" scenario.

\subsubsection{Multi-Doppelganger Mining}
Doppelganger Mining \cite{dm} is a hard class mining method, proposed for the "full softmax" classifier-based face representation learning scenario. Its main idea is to maintain a list with the most similar persons for each person in the training set. These most similar persons are called "doppelgangers", and each person in the dataset has exactly one doppelganger in the list. Doppelgangers are updated at each training iteration, using softmax classification scores of non-target classes: class with the largest score becomes a new doppelganger for a class, which exemplar is classified. At the mini-batch generation stage, a doppelganger list is used to sample pairs of similar classes together.

Doppelganger Mining has demonstrated the ability to improve face recognition models, but it is based on the assumption that all training classes are simultaneously presented in the softmax classifier at each training iteration. It is true for "full softmax"-based training, but false for models, based on "sampled softmax". In the latter case, only a subset of classes is used in the classifier at each iteration, so there is no guarantee that the most similar class ("global" doppelganger) is presented in the current classifier. As the model is forced to update the doppelganger list at each iteration, it may rewrite "global" doppelganger class in the list with a "local" one, by searching only in a subset of currently sampled classes. When the total number of classes in the dataset is large, and the number of sampled classes is small, there is a high chance, that doppelganger list will be filled with mostly "local" doppelgangers, missing the "global" ones, which are more preferable.

To prevent it, we propose Multi-Doppelganger Mining - a modification of Doppelganger Mining, adapted for the usage with "sampled softmax" models like Prototype Memory. With Multi-Doppelganger Mining, each class in the training dataset has a set of doppelgangers in the doppelganger list (instead of just one doppelganger per class as in the original version). Doppelgangers are added to this set using classifier scores in "sampled softmax": non-target class with the largest classification score is added as a new doppelganger to the set. If a class in the doppelganger set is presented in the current "sampled softmax" classifier, but does not get the largest classification score, - it is removed from the doppelganger set. At the stage of mini-batch generation, the doppelganger class is selected randomly from the doppelganger set. 

\subsubsection{Hardness-aware Example Mining}
In the "group-based classes-then-images" mini-batch generation strategy, when the classes are selected (randomly or using hard class mining algorithm), there is a need to select examples of these classes to be used in the mini-batch. While random sampling could be performed successfully in most cases, sometimes the number of examples per class is very large, and only a small portion of them possess enough utility for the training process.

To find the most useful examples and put them in the mini-batches more frequently, we propose a hardness-aware example mining method. Its main idea is to use cosine similarity scores, which are calculated between exemplar embeddings and class prototypes in the process of softmax-based classification, to measure the "hardness" of each example in the dataset, and then use these hardness values in the mini-batch generation stage to sample hard examples more frequently. Calculated hardness values are kept between training iterations and rewritten every time, when the example is classified again. Hardness values are calculated using:

\begin{equation}
	H_i = 1 - \cos\theta_{y_{i}},
\end{equation}
where $y_{i}$ is the ground-truth class for exemplar $i$, and $\theta_{y_{i}}$ is an angle between exemplar embedding and class prototype. Hardness values are initialized with $H_i = 2.0$. To control the amount of hard examples in the mini-batch, hyperparameter $h$ (hardness ratio) is used. It determines the proportion of mini-batch examples, which are sampled using hardness values instead of random sampling. When $h=0.0$, it means that all examples are sampled randomly, $h=1.0$ means all examples are sampled using their hardness values, $h=0.5$ means half of the mini-batch is sampled using hardness, and half of the mini-batch - randomly.

\subsection{Prototype Memory Knowledge Distillation}

Knowledge distillation \cite{hinton2015distilling, gou2021knowledge} is a useful method of transferring knowledge from large teacher models to the smaller student models. There are several knowledge distillation methods, suitable for face recognition models \cite{svitov2020margindistillation, wu2020learning, nekhaev2019margin}, but they are not adapted to the Prototype Memory approach. For example, to initialize the training, \cite{wu2020learning} needs class centers for each class, calculated by teacher network. They could either be taken from the classifier weight matrix, or computed by averaging all teacher embeddings of each class in the dataset. This method requires memory space for prototypes of all dataset classes, thus limiting the ability of Prototype Memory to scale to the unlimited dataset sizes.

Here we propose Prototype Memory Knowledge Distillation (PMKD) - an approach of Prototype Memory-suitable knowledge distillation. Its essence is to calculate embeddings of the training dataset images with a teacher network (it could be done offline or online, in parallel with the training of the student, on a dataset of any size), and use these embeddings instead of the produced by the student encoder to generate prototypes for the Prototype Memory.

\section{Experiments}

In this section we perform experiments with different Prototype Memory hyperparameters. We demonstrate the ability of Prototype Memory to solve the problem of prototype obsolescence. We also compare the performance of the proposed model with different "sampled softmax"-based alternatives and evaluate the effectiveness of the proposed hard example mining and knowledge distillation methods. Finally, we provide the comparison of the proposed Prototype Memory model with the state-of-the-art results on popular face recognition benchmarks.

\subsection{Implementation Details}

\subsubsection{Datasets}
For training we used Glint360k \cite{an2020partial} dataset, containing $360,232$ persons and $17M$ images, its cleaned variant Glint360k-M, containing $356,892$ persons (we found identity duplicates and merged them) and $17M$ images, and also Glint360k-R with $353,658$ persons and $16.3M$ images, where we removed identities, overlapping with testing datasets to perform more strict and fair evaluations \cite{inclusion}. For the experiments with prototype obsolescence prevention, we also used cleaned version of MS-Celeb-1M \cite{guo2016ms} dataset, containing $90,382$ persons and $5.2M$ images. For testing we used datasets: LFW \cite{lfw}, CFP-FP \cite{cfpfp}, AgeDB-30 \cite{agedb}, CALFW \cite{calfw}, CPLFW \cite{cplfw}, TrillionPairs \cite{trillionpairs} and MegaFace, original \cite{kemelmacher2016megaface} and refined (R) version \cite{deng2019arcface}.

\subsubsection{Training Settings}
We use face images of size $112 \times 112$, detected and cropped by RetinaFace detector \cite{deng2020retinaface}, and employ ResNet-34 and ResNet-100 \cite{he2016deep} architectures for the encoder. Final $L_2$-normalized embeddings are of size $D=256$ for ResNet-34 and $D=512$ for ResNet-100 models. For experiments with Prototype Memory hyperparameters we used CosFace \cite{wang2018cosface} loss with $s=64$ and $m=0.4$. The learning rate started from $0.1$ and divided by $10$ at $100k$, $200k$, $250k$, $275k$ iterations finishing at $300k$ iterations. For the experiments with larger networks, we used PM-100: ResNet-100, pre-trained using Prototype Memory ($M=200,000$, $r=0.2$) and CosFace with $s=64$ and $m=0.4$. The learning rate started from $0.1$ and divided by $10$ at $200k$, $400k$, and $500k$ iterations, finishing at $540k$. Mini-batch size is $512$ for all models, mini-batch sampling is group-based iterate-and-shuffle.

\subsubsection{Testing Settings}
For testing we used $L_2$-normalized average embedding of image and its horizontally flipped copy. For testing with the ResNet-100 model on the MegaFace dataset we used $L_2$-normalized concatenated embeddings of image and its horizontally flipped copy. For the metrics we employ verification accuracy on LFW, CFP-FP, AgeDB-30, CALFW and CPLFW datasets. We also perform evaluations on TrillionPairs (identification TPR@FAR=1e-3 and verification TPR@FAR=1e-9) and on MegaFace, original and refined (identification rank-1 and verification TPR@FAR=1e-6).

\subsection{Hyperparameter Effects}

We have performed the experiments with different hyperparameters of Prototype Memory.

\subsubsection{Number of images per class (k) for prototype generation}
We have performed experiments with the number of images per class ($k$) in the mini-batch. We used a ResNet-34 model with memory size of $M=36,000$ and refresh ratio $r=0.2$, trained on Glint360k with a mini-batch of size $512$, and performed the testing on LFW, CFP-FP, AgeDB and Trillion-Pairs. The results are in Table 1.

\begin{table}[h]
	\begin{center}
		\label{table:binsize}
		\begin{tabular}{|c|ccc|cc|}
			\hline
			\multirow{2}{*}{$k$} & \multirow{2}{*}{LFW} & \multirow{2}{*}{CFP-FP} & \multirow{2}{*}{AgeDB} & \multicolumn{2}{|c|}{Trillion-Pairs} \\
			\cline{5-6}
			&&&& Id & Ver\\
			\hline\hline
			3 & 99.70 & \textbf{98.50} & 97.95 & 72.18 & 73.07\\
			4 & \textbf{99.73} & 98.40 & \textbf{98.20} & \textbf{76.03} & \textbf{75.27}\\
			5 & 99.67 & 98.36 & 97.72 & 75.22 & 74.92\\
			\hline
		\end{tabular}
	\end{center}
	\caption{Effects of the number of images per class for prototype generation}
\end{table}

The best results were achieved with $k=4$ images per class. Smaller value of $k$ results in less accurately generated prototypes, larger $k$ results in smaller number of different identities in each mini-batch. With larger mini-batch sizes and with datasets with many images per class it is reasonable to try larger values of $k$. 

\subsubsection{Refresh Ratio}
We have performed experiments with different Prototype Memory refresh ratios ($r$). $r=0.0$ means prototypes in memory are not refreshed, $r=1.0$ - prototypes in memory are completely replaced with new prototypes, $r=0.2$ and $r=0.5$ are other values, providing different levels of the refreshing strength. We used a ResNet-34 model with memory size of $M=36,000$ and $k=4$ images per class in the mini-batch, trained on Glint360k, and performed the testing on LFW, CFP-FP, AgeDB and Trillion-Pairs. The results are in Table 2.

\begin{table}[h]
	\begin{center}
		\label{table:refresh_ratio}
		\begin{tabular}{|c|ccc|cc|}
			\hline
			\multirow{2}{*}{$r$} & \multirow{2}{*}{LFW} & \multirow{2}{*}{CFP-FP} & \multirow{2}{*}{AgeDB} & \multicolumn{2}{|c|}{Trillion-Pairs} \\
			\cline{5-6}
			&&&& Id & Ver\\
			\hline\hline
			0.0 & \textbf{99.73} & \textbf{98.47} & 97.70 & 73.76 & 72.95\\
			0.2 & \textbf{99.73} & 98.40 & \textbf{98.20} & \textbf{76.03} & \textbf{75.27}\\
			0.5 & \textbf{99.73} & 98.36 & 97.37 & 66.15 & 65.80\\
			1.0 & 99.45 & 95.27 & 95.80 & 53.98 & 53.41\\
			\hline
		\end{tabular}
	\end{center}
	\caption{Effects of the refresh ratio ($r$)}
\end{table}

The best results were achieved with $r=0.2$, demonstrating the effectiveness of prototype refreshing. With $r=0.0$ prototypes are getting updates from their class exemplars only in a form of infrequent gradients, leading to a mild form of "prototype obsolescence". Large $r$ values refresh prototypes too strongly, resulting in the loss of valuable information, accumulated during training by the prototypes in memory. $r=0.2$ is a value, achieving the balance between these two situations.

\subsubsection{Memory Size}
We have performed experiments with different memory sizes $M$ for Prototype Memory. We used a ResNet-34 model with memory size ranging from $36,000$ to $288,000$, with other hyperparameters set as $r=0.2$, and $k=4$ images per class in the mini-batch of size $512$. We have trained the models on Glint360k-M, and performed the testing on MegaFace (R) identification benchmark. The results are presented in Fig. 4.

\begin{figure}
	\begin{center}
		\includegraphics[width=1.0\linewidth]{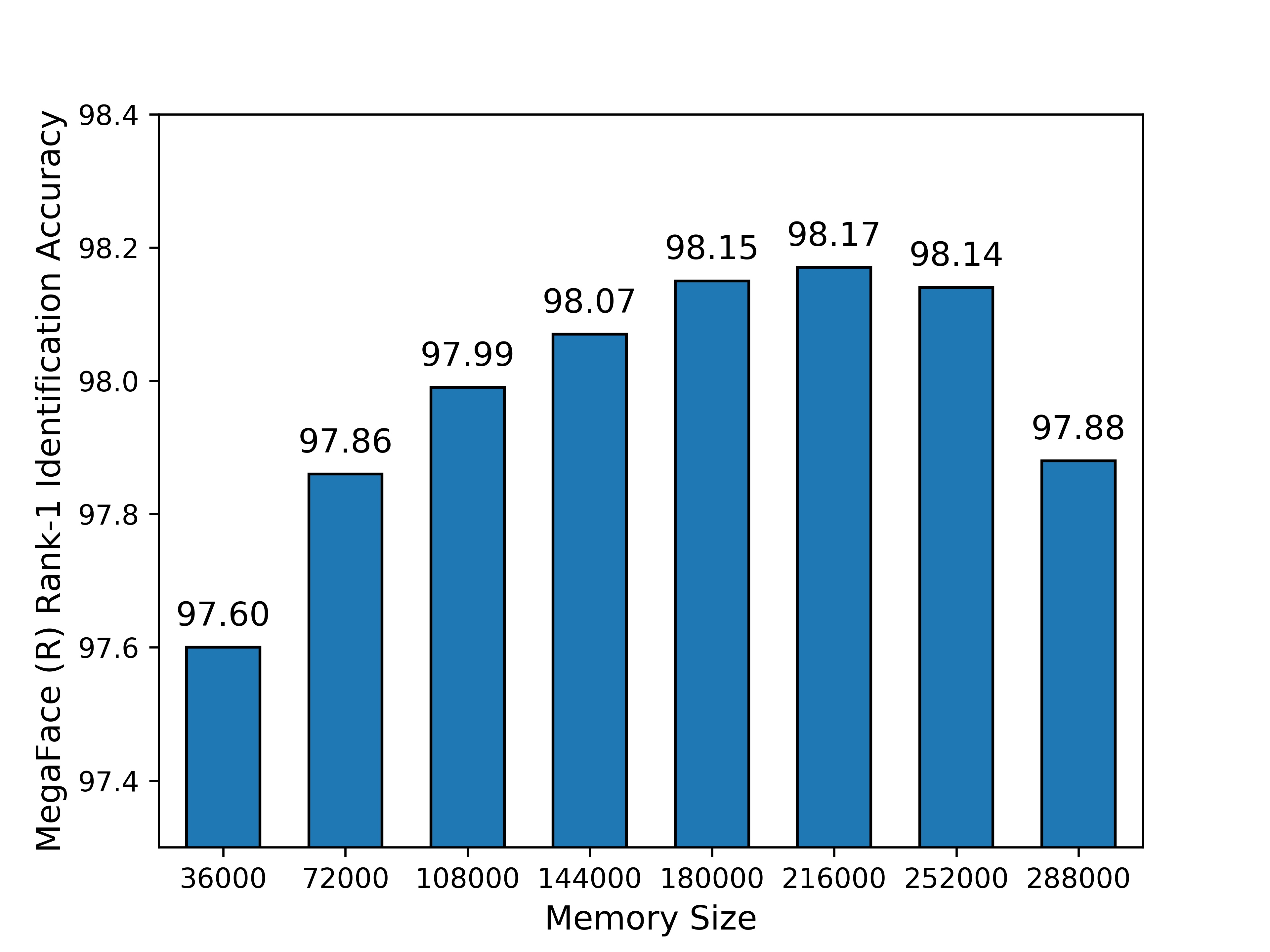}
	\end{center}
	\caption{Effect of the memory size M on MegaFace (R) rank-1 identification accuracy}
	\label{fig:Smirn5}
\end{figure}

The best results were achieved with memory size of $216,000$. With the increase of the memory size from $36,000$ to $216,000$, performance of the model improved. Presence of the sufficient number of prototypes in memory is essential for the construction of informative training signals. Large memory with correctly calculated prototypes provides a comprehensive model of the class distribution in the embedding space, which is used for accurate calculation of gradients to train precise face recognition models. However, it is hard to maintain the correctness of all prototypes in large-scale memory, when the encoder, which produces face representations, is evolving, and prototypes in the memory are updated with delays and imperfect information.

The decrease of the face recognition accuracy for larger memory sizes in our experiments illustrates the harmful impact of the infrequency of the target class exemplar-based gradient updates. With the large memory sizes, most gradient updates, received by the prototypes in memory, come from the non-target class examples, pushing these prototypes away from them in the embedding space. Target class exemplar-based updates, on the contrary, are rare and thus provide insufficient training signals to the prototypes to keep them in close connection with exemplar embeddings. Prototype refreshing is helpful for this situation, but it is also dependent on the presence of target class examples in the mini-batch, and for large memory sizes its effect is limited. Therefore, to achieve good results, memory size should be large enough, but at the same time limited to a certain size.

\subsection{Experiments with Prototype Obsolescence}

To measure the effect of prototype obsolescence, we calculated average distances between class prototypes and class centers for $100$ longest unsampled classes for Positive Plus Random Negative (PPRN) model \cite{an2020partial} and Prototype Memory. Class prototypes were taken from the weight matrix in RAM for the PPRN model, and generated using $k=4$ random images of the considered class for Prototype Memory. Class centers were calculated as mean embeddings for the corresponding classes, using all class images, both for PPRN and Prototype Memory. We have performed experiments on the cleaned MS-Celeb-1M dataset with $90,382$ classes and on the Glint-360k-R dataset with $353,658$ classes. We used a softmax size of $36,000$ ($\approx 0.1$ sampling rate) for both PPRN and Prototype Memory, and trained for $50,000$ iterations. We used the ResNet-34 model as the encoder and CosFace with $s=64$ and $m=0.4$ for the loss function. Prototype Memory used $r=0.2$ and $k=4$. Results of the experiments are presented on Fig. 5.

\begin{figure}
	\begin{center}
		\includegraphics[width=1.0\linewidth]{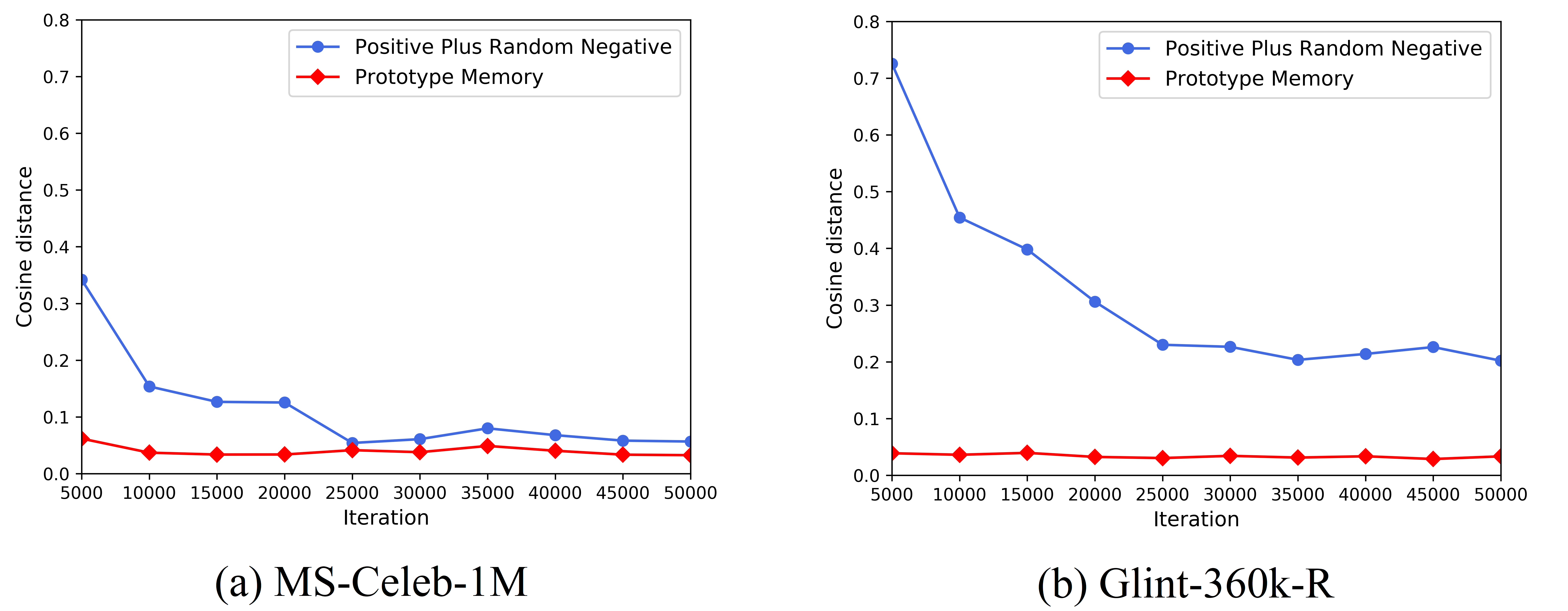}
	\end{center}
	\caption{Average cosine distances between class prototypes and class centers for 100 longest unsampled classes a) on MS-Celeb-1M dataset, without prototype obsolescence problem, b) on Glint-360k-R dataset, with prototype obsolescence problem}
	\label{fig:Smirn4}
\end{figure}

On MS-Celeb-1M dataset (Fig. 5a), where the number of classes is moderate, prototype obsolescence is not present. After a few iterations, class prototypes for the longest unsampled classes move close to the class centers, both for PPRN and Prototype Memory. For the Glint-360k-R dataset (Fig. 5b), however, even after a large number of iterations the distance between class prototypes and class centers for PPRN remains considerably large. This discrepancy between prototypes and class centers is a variant of the problem of prototype obsolescence. It resulted in incorrect training signals, directing the updates in wrong directions. For the datasets of larger size this effect will be even more severe.

Unlike PPRN, the distance between class centers and prototypes for Prototype Memory remains insignificant both for MS-Celeb-1M and Glint-360k-R datasets. Thus we demonstrate that prototype obsolescence is not present in Prototype Memory. With the help of prototype refreshing algorithm, and due to the disposal and subsequent re-generation of the oldest prototypes, and by the means of the limited size of memory module, prototypes in memory remain up-to-date and close to class centers at each training iteration, resulting in more accurate training signals and ability to train face recognition models regardless of the dataset size. 

\subsection{Comparison with Other "Sampled Softmax"-based Methods}

We have performed the comparison with other "sampled softmax"-based methods - Positive Plus Random Negative (PPRN) \cite{an2020partial} and D-Softmax-K \cite{he2020softmax}. We used a ResNet-34 model with memory size of $36,000$ ($\approx 0.1$ sampling rate), $r=0.2$ and $k=4$ images per class. For PPRN we used a sampling rate of $0.1$ and CosFace with $s=64$ and $m=0.4$. For D-Softmax-K we used $d=0.9$, $s=64$ and sampling rate of $0.1$. Training was performed on Glint360k-M, testing on MegaFace, original and cleaned variants (the latter is more reliable). Results of the experiments are presented in Table 3.

\begin{table}[h]
	\begin{center}
		\label{table:samplingbased}
		\begin{tabular}{|c|cc|cc|}
			\hline
			\multirow{2}{*}{Method} & \multicolumn{2}{|c|}{MegaFace} & \multicolumn{2}{|c|}{MegaFace (R)} \\
			\cline{2-5}
			& Id & Ver & Id & Ver\\
			\hline\hline
			PPRN \cite{an2020partial} & 80.43 & 96.19 & 97.33 & 97.54\\
			D-Softmax-K \cite{he2020softmax} & 80.94 & \textbf{96.25} & 97.53 & 97.60\\
			\hline
			Prototype Memory & \textbf{81.48} & 96.02 & \textbf{97.60} & \textbf{98.25}\\
			\hline
		\end{tabular}
	\end{center}
	\caption{Comparison with other "sampled softmax"-based methods}
\end{table}

Prototype Memory outperformed other sampling-based methods in terms of face recognition accuracy. These results prove the effectiveness of the proposed model for learning face representations on large-scale training datasets.

Besides being accurate, Prototype Memory is also more memory-efficient than other softmax-based methods. In Table 4 we compare GPU and Non-GPU memory requirements of Prototype Memory, "full softmax"-based training, PPRN and D-Softmax-K. For the case of "full softmax", full weight matrix of size $D \times N$ should be placed in GPU memory, where $D$ is embedding size and $N$ is total number of classes in the training dataset. There is no need for additional non-GPU memory. For PPRN and D-Softmax-K, only a small part of the full weight matrix is placed in GPU memory, however there is a need to keep the full weight matrix in non-GPU memory. The size of required GPU memory is $D \times M$, where $M < N$ is the number of classes in sampled softmax. The size of required non-GPU memory is $D \times N$.

For the case of Prototype Memory, there is no need to keep a full weight matrix in non-GPU memory, only $D \times M$ of GPU memory is required. Prototype Memory provides a memory-efficient way of training face representations. Memory requirements of the proposed model are independent from the number of classes $N$, making it especially useful for the training on datasets with large numbers of persons.

\begin{table}[h]
	\begin{center}
		\label{table:samplingbasedmemory}
		\begin{tabular}{|c|c|c|}
			\hline
			Method & GPU Memory & Non-GPU Memory\\
			\hline\hline
			Full Softmax & $D \times N$ & -\\
			PPRN \cite{an2020partial} & $D \times M$ & $D \times N$\\
			D-Softmax-K \cite{he2020softmax} & $D \times M$ & $D \times N$\\
			\hline
			Prototype Memory & $D \times M$ & -\\
			\hline
		\end{tabular}
	\end{center}
	\caption{Memory requirements of softmax-based methods, where N is total number of classes in the dataset, M is the size of sampled softmax (M < N), and D is the size of embedding}
\end{table}

Besides being memory-efficient, Prototype Memory is also more computationally efficient than other softmax-based methods. In Table 5 we compare average computations times of main (and most costly) operations in these methods: cosine similarity calculation between embeddings in the mini-batch and prototypes (weights of classes) in the classifier, prototype generation (for Prototype Memory) and RAM-to-GPU transfer of class weights (for other "sampled softmax"-based methods). Computations are performed on 6 NVIDIA GTX 1080 Ti GPUs, with a mini-batch of size $128$ and embedding (and also prototype) size of $D=256$, using parallel acceleration strategy, described in \cite{deng2019arcface}. Total number of classes is set to $1$ million ($1M$), numbers of sampled classes are set to $1M$, $500k$, $200k$, $100k$ and $50k$. Number of images per class for Prototype Memory is $k=4$. Computation times are averaged over multiple iterations (for fair comparisons, for Prototype Memory we start the measurements after the memory is filled to its limits), and given in milliseconds.

\begin{table}[h]
	\begin{center}
		\label{table:computations}
		\begin{tabular}{|c|ccc|}
			\hline
			\multirow{2}{*}{Method} & \multicolumn{3}{|c|}{Average computation time (ms)} \\
			\cline{2-4}
			& Cosine Sim. & Prototypes & RAM-to-GPU\\
			\hline\hline
			Full Softmax ($1M$) & $139$ & $-$ & $-$\\
			\hline
			PPRN ($1M$ / $1M$) \cite{an2020partial} & $139$ & $-$ & $107$\\
			DS-K ($1M$ / $1M$) \cite{he2020softmax} & $139$ & $-$ & $107$\\
			PM ($1M$ / $1M$) (our) & $139$ & $0.5$ & $-$\\
			\hline
			PPRN ($500k$ / $1M$) & $69$ & $-$ & $54$\\
			DS-K ($500k$ / $1M$) & $69$ & $-$ & $54$\\
			PM ($500k$ / $1M$) & $69$ & $0.5$ & $-$\\
			\hline
			PPRN ($200k$ / $1M$) & $27$ & $-$ & $25$\\
			DS-K ($200k$ / $1M$) & $27$ & $-$ & $25$\\
			PM ($200k$ / $1M$) & $27$ & $0.5$ & $-$\\
			\hline
			PPRN ($100k$ / $1M$) & $14$ & $-$ & $12$\\
			DS-K ($100k$ / $1M$) & $14$ & $-$ & $12$\\
			PM ($100k$ / $1M$) & $14$ & $0.5$ & $-$\\
			\hline
			PPRN ($50k$ / $1M$) & $7$ & $-$ & $7$\\
			DS-K ($50k$ / $1M$) & $7$ & $-$ & $7$\\
			PM ($50k$ / $1M$) & $7$ & $0.5$ & $-$\\
			\hline
		\end{tabular}
	\end{center}
	\caption{Comparison of average computation times (in milliseconds) of main operations of different softmax-based models. Prototype Memory (PM) is compared to Full Softmax, D-Softmax-K (DS-K) \cite{he2020softmax} and Positive Plus Randomly Negative (PPRN) \cite{an2020partial} models. Numbers are given for cosine similarity scores computation, prototype generation and RAM-to-GPU weights transfer. Values in brackets represent numbers of sampled and total classes. Mini-batch size is $128$, prototype size is $256$}
\end{table}

As we can see from Table 5, "sampled softmax"-based methods are more efficient than full softmax, when the number of sampled classes is small. However, D-Softmax-K and PPRN methods include the costly RAM-to-GPU class weights transfer operation. When the number of sampled classes is larger, it introduces significant additional computational overhead. For some cases this operation could be optimized (for example, when some class weights are already on GPU since previous training iteration), but in general case it still remains as an unavoidable nuisance. On the contrary, Prototype Memory keeps all computations entirely on GPU and is released from this drawback. On the other hand, Prototype Memory introduces the operation of prototype generation, which also uses some extra computation, but is independent of the number of classes and much faster.

\subsection{Experiments with Multi-Doppelganger Mining and Hardness-aware example mining}

To evaluate the effectiveness of the proposed Multi-Doppelganger Mining and Hardness-aware example mining algorithms, we have performed experiments with PM-100 model, pre-trained on Glint360k-R for $540,000$ iterations using Prototype Memory with $M=200,000$, $r=0.2$, $k=4$ and CosFace with $s=64$ and $m=0.4$ as loss function. This model was fine-tuned for $55,000$ more iterations, with and without hard example mining methods applied.

For the case of usual training, we used composite mini-batch \cite{cmb}, containing $128$ images, sampled with "group-based iterate-and-shuffle" strategy, and $384$ images, sampled with "group-based classes-then-images" strategy. We also used $m=0.5$ as margin value in CosFace.

For the case of hard example mining, we used composite mini-batch, containing $128$ images, sampled with "group-based iterate-and-shuffle" strategy, and $384$ images, sampled using a combination of Multi-Doppelganger Mining and Hardness-aware example mining, with $h=0.5$, $12$ classes sampled at random and $84$ classes sampled using doppelgangers. Margin value $m=0.5$ was used in CosFace. The results of the experiments are presented in Table 6. Multi-Doppelganger Mining and Hardness-aware example mining methods improve the results on MegaFace benchmark. 

\begin{table}[h]
	\begin{center}
		\label{table:mdm}
		\begin{tabular}{|c|cc|cc|}
			\hline
			\multirow{2}{*}{Method} & \multicolumn{2}{|c|}{MegaFace} & \multicolumn{2}{|c|}{MegaFace (R)} \\
			\cline{2-5}
			& Id & Ver & Id & Ver\\
			\hline\hline
			Prototype Memory & \textbf{82.14} & 96.51 & 98.34 & 97.84\\
			Prototype Memory + MDM-HEM & 81.88 & \textbf{96.67} & \textbf{98.47} & \textbf{98.37}\\
			\hline
		\end{tabular}
	\end{center}
	\caption{Experiments with Multi-Doppelganger Mining (MDM) and Hardness-aware example mining (HEM)}
\end{table}

\subsection{Experiments with Knowledge Distillation}

We have performed the experiments with Prototype Memory Knowledge Distillation (PMKD). We used embeddings, calculated by EfficientPolyFace \cite{EPF} network, as teacher embeddings and a ResNet-34 model with memory size of $36,000$, $r=0.2$ and $k=4$ images per class as a student. Training was performed on Glint360k-M, testing on MegaFace, original and cleaned variants (the latter is more reliable). Results are in Table 7. Proposed knowledge distillation method achieved substantial improvement over the student model.

\begin{table}[h]
	\begin{center}
		\label{table:kd}
		\begin{tabular}{|c|cc|cc|}
			\hline
			\multirow{2}{*}{Method} & \multicolumn{2}{|c|}{MegaFace} & \multicolumn{2}{|c|}{MegaFace (R)} \\
			\cline{2-5}
			& Id & Ver & Id & Ver\\
			\hline\hline
			\textit{EPF (Teacher)} \cite{EPF} & \textit{83.16} & \textit{96.94} & \textit{99.12} & \textit{98.54}\\
			\hline
			R-34 (Baseline) & 81.48 & 96.02 & 97.60 & 98.25\\
			R-34 + PMKD (Student) & \textbf{81.66} & \textbf{96.94} & \textbf{98.32} & \textbf{98.32}\\
			\hline
		\end{tabular}
	\end{center}
	\caption{Experiments with Prototype Memory Knowledge Distillation}
\end{table}

\begin{table*}[h]
	\begin{center}
		\label{table:sota-cp}
		\begin{tabular}{|c|c|c|c|c|}
			\hline
			Method & CFP-FP & AgeDB-30 & CALFW & CPLFW \\
			\hline
			\multicolumn{5}{|c|}{With possible train / test identity overlap:} \\
			\hline
			Affinity Loss \cite{hayat2019gaussian} & 96.00 & 95.90 & - & - \\
			Khan et al \cite{khan2019striking} & 97.00 & 94.40 & - & - \\
			D-Softmax \cite{he2020softmax} & 93.07 & 97.30 & - & - \\
			ISL \cite{zhou2019improved} & - & - & 92.72 & 86.13 \\
			ShrinkTeaNet-MFNR \cite{stn} & 95.14 & 97.63 & - & - \\
			RIM \cite{kong2019cross} & - & - & 94.30 & 90.80 \\
			HPDA \cite{Wang_2020_CVPR} & - & - & 95.90 & 92.35 \\
			SFace \cite{zhong2021sface} & - & - & 96.07 & 93.28 \\
			DB \cite{cao2020domain} & - & 97.90 & 96.08 & 92.63 \\
			MultiFace \cite{xu2021multiface} & - & - & 96.08 & 93.32 \\
			QAMFace \cite{zhao2020qamface} & - & - & 96.10 & 92.85 \\
			OL \cite{yang2020orthogonality} & - & - & 96.12 & 92.65 \\
			LATSE \cite{huang2020information} & - & - & 96.20 & 93.48 \\
			AFRN \cite{Kang_2019_ICCV} & 95.56 & 95.35 & 96.30 & 93.48 \\
			SRANet \cite{ling2019self} & 95.60 & 98.47 & - & - \\
			ACNN \cite{ling2019attention} & 95.85 & 98.57 & - & - \\
			MBFN \cite{zhang2019mbfn} & 96.75 & \textbf{98.96} & - & - \\
			CurricularFace \cite{curricularface} & 98.37 & 98.32 & 96.20 & 93.13 \\
			ProbFace \cite{chen2021reliable} & 98.41 & 97.90 & 96.02 & 93.53 \\
			MagFace \cite{meng2021magface} & 98.46 & 98.17 & 96.15 & 92.87 \\
			DiscFace \cite{kim2020discface} & 98.54 & 98.35 & 96.15 & 92.87 \\
			GroupFace \cite{groupface} & 98.63 & 98.28 & 96.20 & 93.17 \\
			BroadFace \cite{kim2020broadface} & 98.63 & 98.38 & 96.20 & 93.17 \\
			DAIL \cite{wang2021dail} & 98.70 & 98.20 & - & - \\
			DDL \cite{huang2020distribution} & 99.06 & - & - & 94.20 \\
			Sub-center ArcFace \cite{deng2020sub} & 99.11 & 98.35 & - & - \\
			Partial FC (r=0.1) \cite{an2020partial} & 99.33 & 98.57 & 96.20 & 94.83 \\
			Partial FC (r=1.0) \cite{an2020partial} & 99.33 & 98.55 & 96.21 & 94.78 \\
			BioMetricNet \cite{ali2020biometricnet} & 99.35 & 96.12 & \textbf{97.07} & \textbf{95.60} \\
			RetinaFace + ArcFace \cite{deng2019retinaface} & \textbf{99.49} & 98.60 & - & - \\
			\hline
			\multicolumn{5}{|c|}{Without train / test identity overlap:} \\
			\hline
			Co-Mining (MsCeleb-R) \cite{Wang_2019_ICCV} & 93.32 & 95.80 & 93.28 & 85.70 \\
			MTLFace \cite{huang2021ageinvariant} & - & 96.23 & 95.62 & - \\
			NPCFace \cite{zeng2020npcface} & 95.09 & 97.77 & 95.60 & 89.42 \\
			SST (AM-Softmax) \cite{du2020semi} & 95.10 & 97.20 & 94.62 & 88.35 \\
			MV-AM-Softmax-a (0.2) \cite{wang2020mis} & 95.30 & 98.00 & 95.63 & 89.19 \\
			Search-Softmax \cite{wang2020loss} & 95.64 & 97.75 & 95.40 & 89.50 \\
			MV-AM-Softmax-f (0.25) \cite{wang2020mis} & 95.70 & 98.05 & 95.45 & 89.69 \\
			Co-Mining (VGGFace2-R)) \cite{Wang_2019_ICCV} & 95.87 & 94.05 & 91.06 & 87.31 \\
			LMFA + TDN \cite{wang2019deep} & 96.86 & 97.92 & - & - \\
			MTCNN + ArcFace \cite{deng2019arcface} & 98.37 & 98.15 & 95.45 & 92.08 \\		
			\hline
			Prototype Memory + CosFace + MDM-HEM & \textbf{99.23} & 98.28 & \textbf{95.97} & \textbf{94.98}\\
			Prototype Memory + D-Softmax & 99.19 & \textbf{98.37} & 95.95 & 94.72\\
			\hline
		\end{tabular}
	\end{center}
	\caption{Comparison with state-of-the-art on CFP-FP, AgeDB-30, CALFW and CPLFW}
\end{table*}

\subsection{Comparison with State-of-the-Art}

We have compared Prototype Memory to the state-of-the-art models from the literature. To perform fair evaluation \cite{inclusion}, we used strict protocol with removed test / train identity overlaps. For the evaluation we used the PM-100 model, fine-tuned with Multi-Doppelganger Mining and Hardness-aware example mining, as mentioned above.

We also used PM-100 to fine-tune another model, containing Prototype Memory $M=200,000$, $k=4$, $r=0.2$, and using D-Softmax \cite{he2020softmax} loss function with $D=0.9$ and $s=64$. Fine-tuning was performed on Glint-360k-R dataset for $55,000$ iterations with mini-batch of $512$, containing $128$ images, sampled with "group-based iterate-and-shuffle" strategy, and $384$ images, sampled with "group-based classes-then-images" strategy. We evaluate both these models on a variety of face recognition benchmarks.

\subsubsection{CFP-FP, AgeDB-30, CALFW and CPLFW}
We have compared Prototype Memory models to the state-of-the-art on CFP-FP \cite{cfpfp} and CPLFW \cite{cplfw} testing datasets, containing large pose variations. We also used AgeDB-30 \cite{agedb} and CALFW \cite{calfw} testing datasets, containing large age variations. Results are in Table 8. Models, which explicitly stated, that they have removed train / test identity overlaps (i.e. use strict evaluation protocol), are placed separately from other models, as the training with train / test overlaps could provide unfair and unreliable test results.

For the strict evaluation protocol, Prototype Memory models achieved state-of-the-art results on all four datasets. Prototype Memory model, trained using CosFace with hard example mining, performed better than Prototype Memory model with D-Softmax on all datasets except AgeDB-30, where the latter model achieved slightly superior results.

\subsubsection{MegaFace}
We have compared Prototype Memory models to the state-of-the-art on large-scale MegaFace benchmark, in identification and verification scenarios, original and cleaned versions. Results are presented in Table 9. For the strict evaluation protocol, and among the models, trained using publicly available face datasets (with the exception of WebFace42M, which is a much larger training dataset), Prototype Memory models achieve state-of-the-art results. Prototype Memory model with D-Softmax outperforms Prototype Memory model with CosFace and hard example mining on verification tests and also on identification test for noisy version of MegaFace. On the identification test of cleaned MegaFace, CosFace-based model achieves better results. These results prove the ability of the models, trained with Prototype Memory, to perform accurate face recognition in large-scale scenarios.

Even better results could be achieved with larger training datasets\cite{zhu2021webface260m}, better encoder architectures \cite{khan2021transformers}, data augmentation \cite{buslaev2020albumentations} and other methods \cite{sanakoyeu2021improving}, providing models with the robustness to large age \cite{zhao2020towards} and pose \cite{ahmed2020frontiers} variations and ability to overcome the problems of racial bias \cite{sixta2020fairface}, domain imbalance \cite{cao2020domain} and bad image quality \cite{luevano2021study}.

\begin{table}
	\begin{center}
		\label{table:sota-mf}
		\begin{tabular}{|c|cc|cc|}
			\hline
			\multirow{2}{*}{Method} & \multicolumn{2}{|c|}{MegaFace} & \multicolumn{2}{|c|}{MegaFace (R)} \\
			\cline{2-5}
			& Id & Ver & Id & Ver\\
			\hline
			\multicolumn{5}{|c|}{With possible train / test identity overlap:} \\
			\hline
			SphereFace \cite{liu2017sphereface} & 72.73 & 85.56 & - & - \\
			SphereFace+ \cite{liu2018learning} & 73.03 & - & - & - \\
			CCL with AAM \cite{qi2018face} & 73.74 & - & - & - \\
			AM-Softmax \cite{wang2018additive} & 75.23 & 87.06 & - & - \\
			Ring Loss \cite{zheng2018ring} & 75.22 & - & - & - \\
			EEL \cite{wu2020learning} & 76.40 & 90.70 & - & - \\
			ShuffleFaceNet \cite{Martindez-Diaz_2019_ICCV_Workshops} & 77.40 & 93.00 & - & - \\
			Fair Loss \cite{Liu_2019_ICCV} & 77.45 & 92.87 & - & - \\
			Wang et al \cite{wang2017multi} & 77.74 & 79.24 & - & - \\
			BUPT-CBFace \cite{zhang2020classb} & 79.57 & 95.20 & - & - \\
			CLMLE \cite{huang2019deep} & 79.68 & 91.85 & - & - \\
			UniformFace \cite{duan2019uniformface} & 79.98 & 95.36 & - & - \\
			Marginal Loss \cite{marginal2017} & 80.28 & 92.64 & - & - \\
			m-LinCos-Softmax \cite{ou2020lincos} & - & - & 88.69 & 91.22 \\
			AdaptiveFace \cite{liu2019adaptiveface} & - & - & 95.02 & 95.61 \\
			DB \cite{cao2020domain} & - & - & 96.35 & 96.56 \\
			ATAM \cite{iranmanesh2020attribute} & - & - & 96.51 & 97.14 \\
			D-Softmax \cite{he2020softmax} & - & - & 97.02 & - \\
			NPT-Loss \cite{khalid2021nptloss} & - & - & 97.67 & 98.54 \\
			MBConv \cite{Lyu_2019_ICCV_Workshops} & 80.29 & 96.47 & 97.80 & 97.94 \\
			AirFace \cite{Li_2019_ICCV_Workshops} & 80.80 & 96.52 & 98.04 & 97.93 \\
			DUL$_{rgs}$ \cite{chang2020data} & - & - & 98.12 & 98.78 \\
			SRANet \cite{ling2019self} & - & - & 98.34 & 98.38 \\
			ACNN \cite{ling2019attention} & - & - & 98.35 & 98.42 \\
			OL \cite{yang2020orthogonality} & 81.05 & 97.16 & 98.37 & 98.80 \\
			IBN-block \cite{qing2018improve} & - & - & 98.43 & - \\
			Circle Loss \cite{sun2020circle} & - & - & 98.50 & - \\
			SFace \cite{zhong2021sface} & 81.15 & 97.11 & 98.50 & 98.61 \\
			IDA-R \cite{liu2021interclass} & - & - & 98.56 & - \\
			DUL$_{cls}$ \cite{chang2020data} & - & - & 98.60 & 98.67 \\
			BA \cite{wei2020balanced} & - & - & 98.62 & - \\
			MultiFace \cite{xu2021multiface} & 80.91 & 97.37 & 98.62 & 98.85 \\
			DiscFace \cite{kim2020discface} & 81.23 & 97.44 & - & - \\
			BroadFace \cite{kim2020broadface} & 81.33 & \textbf{97.56} & 98.70 & 98.95 \\
			CurricularFace \cite{curricularface} & 81.26 & 97.26 & 98.71 & 98.64 \\
			GroupFace \cite{groupface} & 81.31 & 97.35 & 98.74 & 98.79 \\
			DFN-152 \cite{he2020deformable} & 82.11 & - & - & - \\
			IMDb-Face \cite{wang2018devil} & 84.06 & - & - & - \\
			CDP$\textsuperscript{†}$ \cite{zhan2018consensus} & 84.74 & - & - & - \\
			CosFace + DAG$\textsuperscript{†}$ \cite{dabouei2020boosting} & \textbf{85.62} & 97.26 & - & - \\
			Sub-center ArcFace \cite{deng2020sub} & - & - & 98.78 & 98.69 \\
			Partial FC (r=0.1) \cite{an2020partial} & - & - & 98.94 & 99.10 \\
			Partial FC (r=1.0) \cite{an2020partial} & - & - & 99.13 & 98.98 \\
			LATSE \cite{huang2020information} & - & - & \textbf{99.14} & \textbf{99.19} \\
			\hline
			\multicolumn{5}{|c|}{Without train / test identity overlap:} \\
			\hline
			Center Loss \cite{wen2016discriminative} & 65.23 & 76.52 & - & - \\
			AM-Softmax \cite{wang2018additive} & 72.47 & 84.44 & - & - \\
			LightCNN-29 \cite{wu2018light} & 73.75 & 85.13 & - & - \\
			RegularFace \cite{zhao2019regularface} & 75.61 & 91.13 & - & - \\
			COCO \cite{liu2017rethinking} & 76.57 & - & - & - \\
			MTLFace \cite{huang2021ageinvariant} & 77.06 & - & - & - \\
			FAN-Face \cite{yang2020fan} & 78.32 & 92.83 & - & - \\
			DS-Res64 \cite{liu2020face} & 80.81 & - & - & - \\
			SST (AM-Softmax) \cite{du2020semi} & - & - & 96.27 & 96.96 \\
			Search-Softmax \cite{wang2020loss} & - & - & 96.97 & 97.84 \\
			NPCFace \cite{zeng2020npcface} & - & - & 97.75 & 98.07 \\
			CosFace \cite{deng2019arcface} & 80.56 & 96.56 & 97.91 & 97.91 \\
			MV-AM-Softmax-a \cite{wang2020mis} & - & - & 98.00 & 98.31 \\
			ArcFace \cite{deng2019arcface} & 81.03 & 96.98 & 98.35 & 98.48 \\
			LMFA + TDN \cite{wang2019deep} & 81.42 & 95.22 & - & - \\
			CosFace-Private$\textsuperscript{†}$ \cite{wang2018cosface} & \textbf{82.72} & 96.65 & - & - \\
			MC - MegaFace2$\textsuperscript{†}$ \cite{Zhang_2020_CVPR} & - & - & 98.96 & \textbf{98.97} \\
			MC - FaceGraph$\textsuperscript{†}$ \cite{Zhang_2020_CVPR} & - & - & 99.02 & 98.94 \\
			WebFace42M \cite{zhu2021webface260m} & - & - & \textbf{99.11} & - \\
			\hline
			PM + CosFace + MDM-HEM & 81.88 & 96.67 & 98.47 & 98.37 \\
			PM + D-Softmax & 82.26 & \textbf{97.39} & 98.38 & 98.72 \\
			\hline
		\end{tabular}
	\end{center}
	\caption{Evaluation on MegaFace Benchmark, where $\textsuperscript{†}$ denotes the usage of large-scale private training datasets}
\end{table}

\section{Conclusion} In this paper we proposed Prototype Memory - a novel face representation learning model, opening the possibility to train state-of-the-art face recognition architectures on the datasets of any size. Prototype Memory consists of the memory module for storing class prototypes, and algorithms to perform operations with it. Prototypes are generated online, using exemplar embeddings, presented in the current mini-batch. New prototypes are enqueued to the memory. Prototypes in memory are refreshed and kept up-to-date with the state of the encoder. When the memory is filled, the oldest prototypes are dequeued and disposed of. Prototypes in memory are used in the role of classifier weights for softmax-based face representation learning. Prototype memory is useful for preventing the problem of "prototype obsolescence". Like other "sampled softmax"-based models, it is computationally and memory-efficient. We have performed extensive experimental evaluations on popular face recognition benchmarks and proved the effectiveness of the proposed model. We have compared the performance of Prototype Memory to other "sampled softmax"-based models and demonstrated its superiority both in terms of face recognition accuracy and memory efficiency. We proposed Multi-Doppelganger Mining and Hardness-aware example mining methods to improve the training of models, based on Prototype Memory, with the help of the generation of more informative mini-batches. We also described a knowledge distillation method, suitable for using with Prototype Memory, and proved its effectiveness.

\section*{Acknowledgement}

This work was supported by the Analytical Center for the Government of the Russian Federation (IGK 000000D730321P5Q0002), agreement No. 70-2021-00141.

\bibliographystyle{ieeetr}
\bibliography{vs_small}

\begin{IEEEbiography}[{\includegraphics[width=1in,height=1.25in,clip,keepaspectratio]{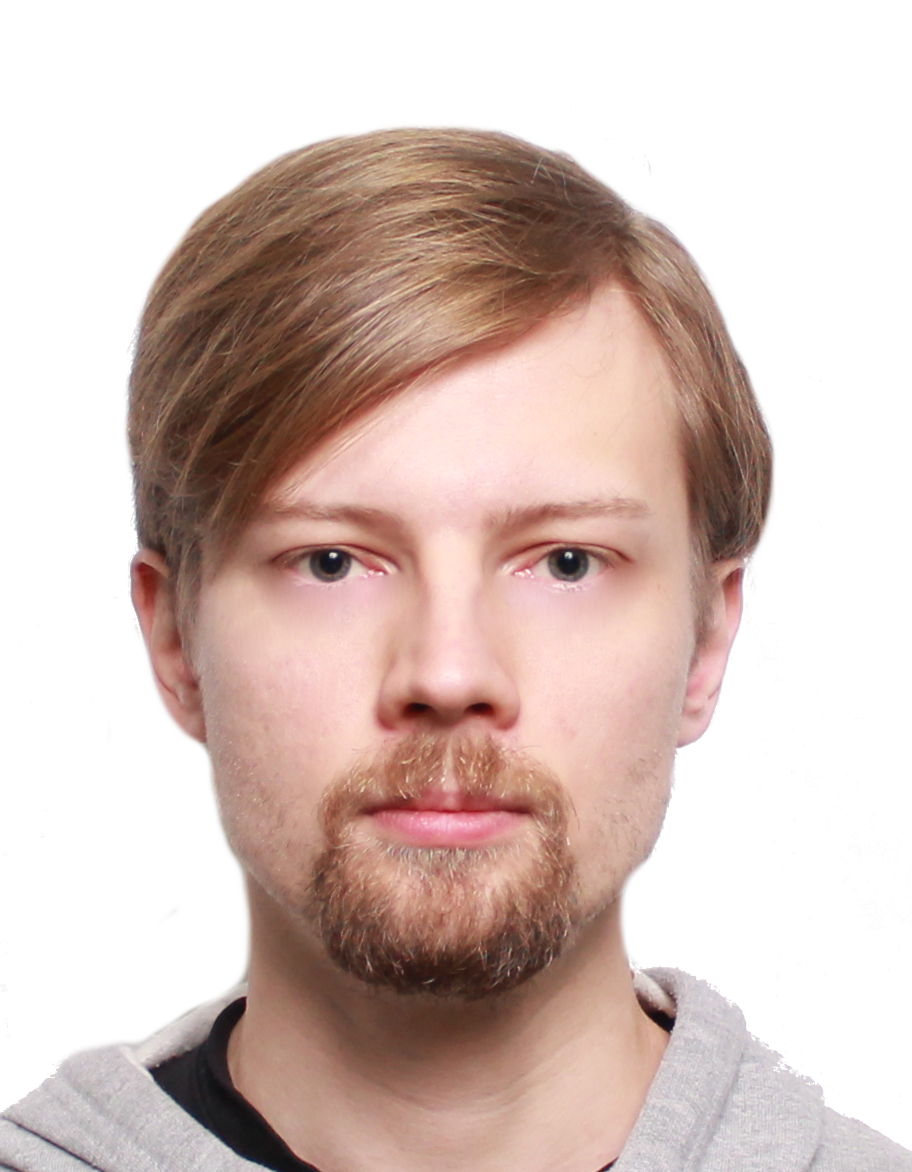}}]{Evgeny Smirnov} received the Specialist Degree in information systems and technology from the ITMO University, Saint Petersburg, Russia, in 2011. He is currently a Senior Researcher with Speech Technology Center (STC). His research interests include computer vision, face recognition, and representation learning.
\end{IEEEbiography}

\begin{IEEEbiography}[{\includegraphics[width=1in,height=1.25in,clip,keepaspectratio]{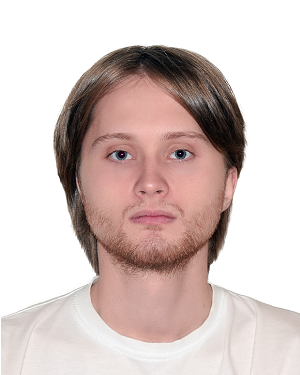}}]{Nikita Garaev} received the B.E. degree from Saint Petersburg Electrotechnical University (ETU "LETI"), Russia, in 2019. He is also received M.S. degree from ITMO University, Russia, in 2021. He now works at the Speech Technology Center. His research interests include computer vision and face recognition.
\end{IEEEbiography}

\begin{IEEEbiography}[{\includegraphics[width=1in,height=1.25in,clip,keepaspectratio]{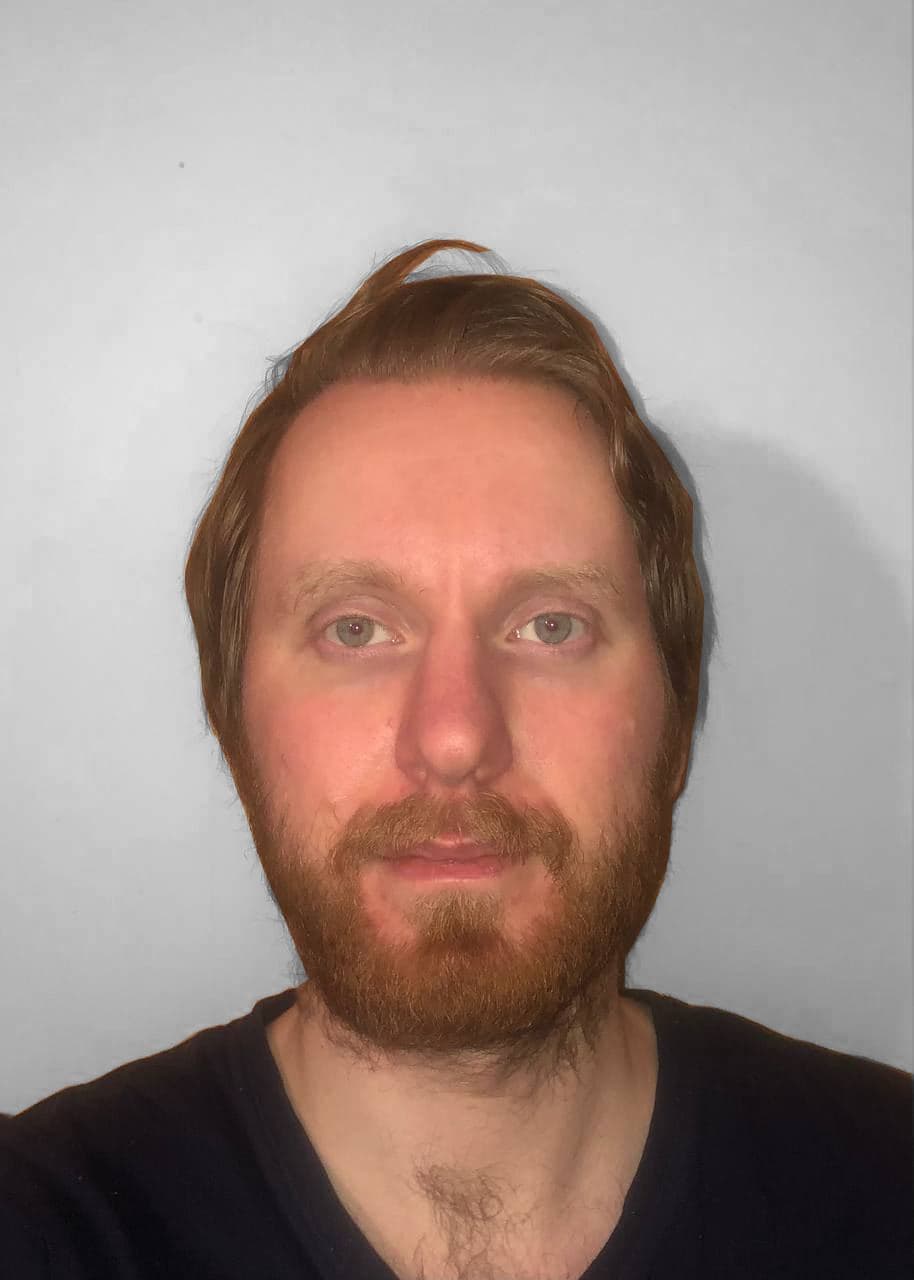}}]{Vasiliy Galyuk} received the B.Sc. degree from the ITMO University, Russia, Saint-Petersburg in 2015, and the master’s degree from the ITMO University, Russia, Saint-Petersburg in 2017. He is currently a Researcher with Speech Technology Center (STC). His research interests include computer vision, image classification, and face detection.  
\end{IEEEbiography}

\begin{IEEEbiography}[{\includegraphics[width=1in,height=1.25in,clip,keepaspectratio]{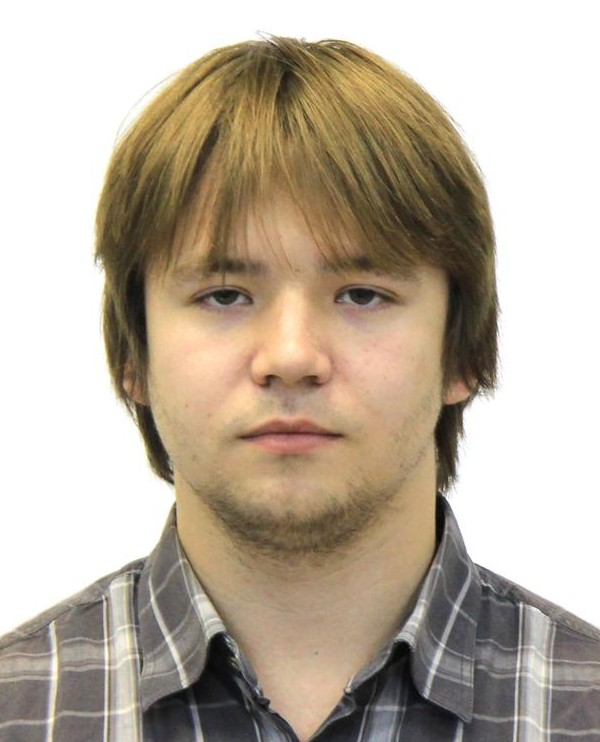}}]{Evgeny Lukyanets} received the Master's Degree in information systems and technology from ITMO University, Saint-Petersburg, Russia, in 2015. He is currently a Developer with ITMO University and Technical Product Lead with Speech Technology Center (STC). His relearch interests include computer vision and face recognition. 
\end{IEEEbiography}

\EOD

\end{document}